\documentclass[letterpaper, 10 pt, conference]{ieeeconf}  
\pdfoutput=1

\IEEEoverridecommandlockouts                              

\usepackage{amsmath} 
\usepackage{amssymb}  
\usepackage{amsmath}
\usepackage{graphicx}
\usepackage{graphbox}
\usepackage[export]{adjustbox}
\usepackage{mathtools}
\usepackage[ruled]{algorithm2e}
\usepackage[pdftex]{pict2e}
\usepackage{tabularx}
\usepackage{tikz}
\usepackage{pgfplots}
\usepackage[colorlinks=true,urlcolor=black]{hyperref}

\hypersetup{
  colorlinks 
}
\hypersetup{linkcolor=black}
\usepackage{url}
\usepackage{float}
\usepackage{subcaption}
\usepackage{multirow}
\usepackage{eurosym}
\usepackage{cleveref}
\usepackage{booktabs} 
\usepackage[inline]{enumitem}
\usepackage{xcolor}
\usepackage{tcolorbox}
\usepackage{pdfpages}
\usepackage{fancyhdr}
\usepackage[%
 toc,              
 acronym,          
 section=chapter,  
 nonumberlist,     
]{glossaries}
\newacronym{MEMS}{MEMS}{Micro-Electro-Mechanical System}
\newacronym{ML}{ML}{Machine Learning}
\newacronym{PCB}{PCB}{Printed Circuit Board}
\newacronym{IMU}{IMU}{Inertial Measurement Unit}
\newacronym{AHRS}{AHRS}{Attitude and Heading Reference Systems}
\newacronym{eKF}{eKF}{extended Kalman Filter}
\newacronym{RMSE}{RMSE}{Root Mean Square Error}
\newacronym{SCADA}{SCADA}{Supervisory Control and Data Acquisition}
\newacronym{BLE}{BLE}{Bluetooth Low Energy}

\newacronym{iot}{IoT}{Internet-of-Things}
\newacronym{vq}{VQ}{Vector Quantization}
\newacronym{rvq}{RVQ}{Residual Vector Quantization}
\newacronym{zfp}{ZFP}{Zero Footprint Compressor}
\newacronym{sz}{SZ}{Scientific Zip}
\newacronym{hpc}{HPC}{High Power Computing}
\newacronym{cr}{CR}{Compression Ratio}
\newacronym{png}{PNG}{Portable Network Graphics}
\newacronym{mp3}{MP3}{MPEG Audio Layer 3}
\newacronym{mse}{MSE}{Mean Squared Error}
\newacronym{mcu}{MCU}{Microcontroller Unit}
\newacronym{soa}{SoA}{State of the Art}
\newacronym{wsn}{WSN}{Wireless Sensor Network}

\usepackage[detect-none,detect-weight=true, detect-family=true]{siunitx}
\title{\LARGE \bf
EdgeCodec: Onboard Lightweight
High Fidelity Neural Compressor
with Residual Vector Quantization
}

\author{Benjamin Hodo$^{1}$, Tommaso Polonelli$^{1}$, Amirhossein Moallemi$^{2}$, Luca Benini$^{1}$, and Michele Magno$^{1}$ \\
\thanks{The experiments were performed within the French-Swiss project MISTERY funded by the French National Research Agency (ANR PRCI grant no. 266157) and the Swiss National Science Foundation (grant no. 200021L 21271). This work has also received funding from the Swiss State Secretariat for Education, Research, and Innovation (SERI) under the SwissChips initiative. Moreover, we truly thank Marco Fariselli for supporting this work.} \thanks{979-8-3315-6578-7/25\$31.00 ©2025 IEEE}
$^{1}$D-ITET, ETH Zürich, Zürich, Switzerland \\
$^{2}$RTDT Laboratories, Switzerland
}%
\begin{document}

\title{\LARGE \bf
{\footnotesize This paper has been accepted for publication at the International Workshop on Advances in Sensors and Interfaces (IWASI), Italy, 2025. ©2025 IEEE. \\
DOI: To be added when available.\vspace{1em}}\\[0.5em]
EdgeCodec: Onboard Lightweight\\
High Fidelity Neural Compressor\\
with Residual Vector Quantization
}
\maketitle
\thispagestyle{fancy}
\fancyhf{}  

\fancyfoot[L]{\footnotesize
\begin{minipage}{0.95\textwidth}
\copyright 2025 IEEE. Personal use of this material is permitted. Permission from IEEE must be obtained for all other uses, in any current or future media, including
reprinting/republishing this material for advertising or promotional purposes, creating new collective works, for resale or redistribution to servers or lists, or reuse of any copyrighted component of this work in other works.
\end{minipage}
}

\begin{abstract}
Data Compression is a staple of data processing and storage. Sending and storing data more efficiently is an open challenge in the \gls{iot}, with devices typically characterized by limited availability of energy and computing power. The problem tackled in this paper is the massive amounts of sensor data collected and sent uncompressed by \gls{iot}-devices. We address this issue by compressing local data using a neural network supplemented with the \gls{rvq} technique. This paper, inspired by lossy neural compressors for audio like Google \textit{Soundstream} and Meta \textit{EnCodec}, proposes \textit{EdgeCodec}: a lightweight lossy neural compressor specifically designed to run at the edge on low-power and resource constrained \glspl{mcu}. \textit{EdgeCodec} processes multi-channel data with a flexible end-to-end learnable pipeline. We evaluate \textit{EdgeCodec} in a real-life challenging use case, namely wind turbine monitoring using a 40-channel barometric sensor. Under the proposed use-case, our \textit{EdgeCodec} reaches a \gls{cr} between 2560 and 10240 that can be varied in real-time to tune the trade-off between compression and reconstruction quality. Executed onboard a low-power \gls{mcu}, \textit{EdgeCodec} achieves an average reconstruction error of 2.54\% over the whole validation set. It requires only \qty{52.6}{\milli\second} to compress the \qty{16}{\kilo b} generated in \qty{1}{\second}. Final results demonstrate how \textit{EdgeCodec} can reduce the energy consumption by 2.9x for the required wireless transmission.
\end{abstract}

\vspace{-5pt}
\section{Introduction}
Data compression is a key step in communication and \gls{iot} applications \cite{jayasankar2021survey}. From stored pictures, streaming video, or audio \cite{al2025data}, the primary motivation is to reduce storage size and communication bandwidth. Lossy compression, commonly used for images and audio, comes with a tradeoff between \gls{cr} and numerical accuracy when reconstructing the original datastream \cite{bozhilov2024systematic}. The field of audio and video has witnessed a significant acceleration with machine-learning (ML) based compression approaches, like Google \textit{SoundStream} \cite{soundstream} or Meta \textit{EnCodec} \cite{meta}, which are competitive with state-of-the-art audio and video codecs like \textit{OPUS} \cite{OPUS} and \textit{EVS} \cite{EVS}. Classical data compression is split into two categories, lossless and lossy \cite{bozhilov2024systematic, jayasankar2021survey}. Lossless compression generally achieves lower \gls{cr} than lossy, but it ensures that the reconstructed object is identical to the original one \cite{al2025data}. A popular lossless compression technique for images is \gls{png}, utilizing delta encoding. On the other hand, lossy compression techniques can achieve very high \gls{cr} by discarding irrelevant or redundant information in the latent space. Lossy compressors can be subdivided into two more subcategories, transform-based and prediction-based techniques \cite{bozhilov2024systematic}. Transform-based techniques use mathematical transformations to go from their original domain to a different one, where redundant information is easier to spot and to be deleted \cite{al2025data,jayasankar2021survey}. An example is \gls{mp3}, transforming audio signals into the frequency domain where high-frequency components can be discarded due to the human ear's sensitivity \cite{OPUS}. The transform-based \gls{zfp}, used as a reference to compare this paper results with the \gls{soa}, utilizes block-based transformations that convert data in lower-magnitude components, while maintaining a strict error-bound \cite{lindstrom2024zfp}. Moreover, \gls{sz} is a prediction-based floating point array compressor that utilizes predictive modeling techniques like linear regression and Lorenzo predictors \cite{SZ}. It estimates the values and encodes only the residual, maintaining user set error-bound and achieving high ($>500$) \gls{cr} \cite{SZ}. \gls{sz} and \gls{zfp} are widely recognized for their contributions to the \gls{soa}, forming a suitable baseline for our paper.
Even though data compression is an active area of research within \gls{iot} \cite{ACM_survey}, its deployment and application remains limited due to the lack of lightweight compression libraries able to run at the edge on resource constrained \glspl{mcu}. The main challenge comes from the limited availability of onboard computation power and memory \cite{fischer2021windnode, von2024tiny} since most compression algorithms are designed for high-end platforms, such as smartphones \cite{soundstream, meta}. Therefore, low-power sensor nodes directly transmit uncompressed data to the cloud, increasing the traffic and the energy used to transfer data wirelessly \cite{polonelli2022aerosense}. Thus, there is a need for lightweight and efficient lossy compression algorithms specifically targeted for low-power \gls{iot} devices, typically featuring \glspl{mcu} with a total memory in the range of \qty{1}{\mega B}. The advantage is twofold: decreasing the transmitted data load to minimize the traffic and increasing the battery lifetime by reducing the energy used to stream.
This paper presents \textit{EdgeCodec}\footnote{\url{github.com/ETH-PBL/EdgeCodec}}, a machine learning-based lossy compression approach specifically designed for low-power and resource-constrained \gls{iot} devices. It is conceptualized on \cite{meta,soundstream} regarding the model structure and the training strategy, but it is heavily optimized for edge processing by pruning unnecessary layers and only using lightweight operators supported by \gls{mcu}'s \gls{ML} accelerator engines. It runs in milliseconds on commercial \glspl{mcu} with a memory requirement below \qty{1.5}{MB} of RAM. To prove \textit{EdgeCodec}'s efficacy, we tested it in a real use case scenarios (wind turbine monitoring), where it reaches a \gls{cr} between 2500 and 10200 with a datastream generated from 40 barometers. A key feature of \textit{EdgeCodec} is the possibility of varying the \gls{cr} in real time. This approach allows the \gls{iot} sensor node to dynamically vary the bitrate depending on the connection status and the available energy. A trade-off between \gls{cr} and compression quality exists; in our field tests, we noticed an average error on the reconstructed data between 2.5\% and 2.9\%, depending on the selected \gls{cr}, values always within the application specification of 3\% discussed in \Cref{sec:dataset}. In detail, this paper's main contributions are:
\begin{enumerate*}[label=(\roman*), font=\itshape]
    \item A novel lightweight lossy compressor for multi-channel time-series data, named \textit{EdgeCodec}, specifically designed for low-power \gls{iot} devices. \textit{EdgeCodec} exploits an unbalanced \gls{ML} autoencoder architecture running in \textit{float16} joint with a \gls{rvq}. Moreover, it features a sample-by-sample variable bitrate. The training approach is end-to-end, including an adversarial model to increase the reconstruction accuracy.
    \item A full deployment at the edge, with a model size of only \qty{58}{\kilo parameter}, requiring less than \qty{1.5}{MB} of RAM. The model compressing \qty{8}{\second} of data, runs in $\sim$\qty{53}{\milli\second} and requires \qty{8}{\milli\joule}. The model is parallelized and deployed on a RISC-V \gls{mcu}.
    \item A comparison with \gls{soa} non-neural compressors, \gls{sz} \cite{SZ} and \gls{zfp} \cite{lindstrom2024zfp} in terms of \gls{cr} and reconstruction error.
\end{enumerate*}

\section{Related Works}
Since the beginning of the internet era, data compression has been a key focus topic \cite{ketshabetswe2021data}; and, even more today, it is a fundamental step to handle the large amount of data generated by the deployed \gls{iot} infrastructure. Today, researchers and large internet companies mainly focus on audio and video real-time compression for streaming and media \cite{jayasankar2021survey}. However, most of the \gls{iot} data generated from low-power sensor nodes have been transmitted on \gls{wsn} uncompressed so far, with obvious consequences of increased data traffic, cost-per-MB of data transported via cellular operators, and scalability issues \cite{al2025data}.  
Common lossy compressors for \gls{iot} target mainly audio (\gls{mp3}) and images (JPEG \cite{polonelli2020energy}), while algorithms targeted for time-series, such as \gls{zfp} and \gls{sz}, require high computational load compared to the \gls{cr}, in general between 15 and 512. Nowadays, with the new advent of Edge computing \cite{kong2022edge}, new lossy compression algorithms can be developed for low power and resource constrained \gls{iot} sensor nodes \cite{kong2022edge}, exploiting \gls{ML} techniques to reach high \gls{cr} ($>1000$) at low latency and low computational effort.
EnCodec is a state-of-the-art real-time, high-fidelity audio codec powered by neural networks \cite{meta}. It employs a streaming encoder-decoder architecture with a quantized latent space, trained end-to-end to achieve efficient and high-quality audio compression \cite{meta}. EnCodec utilizes a single multiscale spectrogram adversary to optimize training, which effectively minimizes artifacts and ensures superior audio reconstruction \cite{gu2024multi}. 
Results consistently demonstrate that EnCodec outperforms baseline methods across all tested conditions, delivering superior quality for both \qty{24}{\kilo\hertz} monophonic and \qty{48}{\kilo\hertz} stereophonic audio. 

SoundStream is a novel neural audio codec designed to efficiently compress speech, music, and general audio, achieving bitrates typically targeted by speech-specific codecs \cite{soundstream}. It features a fully convolutional encoder-decoder architecture combined with a residual vector quantizer\footnote{\url{github.com/lucidrains/vector-quantize-pytorch}}, jointly trained in an end-to-end fashion. 
Particularly interesting for the scope of this paper: a single model can seamlessly operate across variable bitrates ranging from 3 to \qty{18}{kbps}, maintaining near-optimal quality even when compared to models trained at fixed bitrates.
This study builds its foundation on the concept brought by EnCodec and SoundStream, an end-to-end neural compression with residual vector quantization and an adversarial neural network used during training, to develop an Edge computing compatible lossy encoder for multi-channel time-series bitstreams. The proposed approach specifically targets the deployment on low-power and resource constrained \gls{iot} sensor nodes, characterized by a \gls{mcu} with limited amount of onboard memory.

\subsection{Case Study}
\label{sec:casestudy}
For the scope of this paper, we consider a sensor system specifically designed for monitoring and performing predictive maintenance on operating wind turbine blades. 
The Aerosense sensor node \cite{fischer2021windnode} is designed to be thin, minimally intrusive, low-power, and self-sustaining \cite{polonelli2022aerosense}. It transmits wirelessly over \gls{BLE}, making it easy to install and to be produced. For data transmission, the system utilizes a \gls{BLE} downlink with a bandwidth of \qty{1.2}{\mega bps}, consuming up to \qty{178}{\nano\joule} per bit~\cite{fischer2021windnode}. The sensor node combines 40 barometers and 5 differential pressure sensors, as well as 10 acoustic sensors, all mounted on flexible \glspl{PCB} around the wind turbine blade. 
At its maximum speed, Aerosense generated \qty{4.2}{Mbps}, which exceeds the \gls{BLE} downlink bandwidth, while the 40 barometers alone generate \qty{128}{kbps}. To handle this data stream, the energy for the \gls{BLE} wireless transmission is $\sim$50\% of the total power budget \cite{polonelli2022aerosense}.

\subsection{Dataset}
\label{sec:dataset}
This study leverages the open-source dataset presented in~\cite{iabdallah2023}, which was acquired using the Aerosense system~\cite{fischer2021windnode,polonelli2022aerosense}. The experimental setup is designed to simulate the progressive structural degradation of a wind turbine blade by introducing artificial cracks. The dataset is categorized into six damage classes. 
We used this dataset as a case study because of the specific characteristic of the Aerosense system, where the \gls{BLE} power dominates due to the high-speed data stream. Compressing the data would drastically prolong the battery lifetime in this particular scenario. Moreover, the dataset~\cite{iabdallah2023} is already used in the literature to perform damage detection with a data-driven approach and the 40 barometric data \cite{von2024tiny}, which allow up to 3\% data perturbation without affecting damage classification accuracy. We use the 3\% threshold as a training parameter for the scope of this paper, defining the maximum acceptable error between raw data and reconstructed data generated by our \textit{EdgeCodec}. Note that this number may vary for different applications or datasets. 
\vspace{-5pt}
\section{Methodology}

\label{sec:methodo}
%
\Cref{fig:Detailed_model} presents the general architecture of \textit{EdgeCodec}, consisting of three major building blocks:
\begin{enumerate*}[label=(\roman*), font=\itshape]
    \item The encoder, which maps inputs $x$ to a latent representation $E(x)=z$. In our case, the encoder has a native \gls{cr} of 32.
    \item The Residual Vector Quantizer, which further transforms (compresses) the input latent representation to N bits via a finite set of codebooks, where each of them represents a quantized value of each dimension of the latent space.
    \item The decoder, which produces a data reconstruction $\hat{x}$ from the quantized latent space representation.
\end{enumerate*}

\subsubsection{Encoder}

The encoder depicted in \Cref{fig:Detailed_model} consists of four encoder blocks encased by two 1D-convolutional layers. Each encoder block comprises a residual unit and a 1D-convolutional layer paired with a PReLU activation function. Each residual unit consists of two fully bypassed 1D-convolutional layers followed by a PReLU activation function. The encoder has a channel progression of 36 to 9 and a data progression of 800 to 100. Values empirically determined with hyperparameter tuning during training and specifically related to the dataset described in \Cref{sec:dataset}. This yields the previously mentioned native \gls{cr} of 32. The encoder in its current setup has a parameter count of 58'095, requiring \qty{232}{\kilo\byte} of non-volatile memory on the edge \gls{mcu}. While designing the encoder, it was crucial to keep the operations used compatible with \gls{soa} \glspl{mcu}.  In contrast to \textit{SoundStream}, \textit{EdgeCodec} reduces its parameter count by demodulating channels. If one fits Google \textit{SoundStream} with 36 input channels, it amounts to 5'826'249 parameters, which is about 100 times more than \textit{EdgeCodec}s encoder. The complexity of the encoder is easily modifiable due to the compartmentalized structure of blocks and units; for example, one could add residual units to the blocks like Google \textit{SoundStream} or add more encoder blocks and operations in between. This yields a highly versatile architecture.

\subsubsection{Residual Vector Quantizer}
The \gls{rvq} works by quantizing the latent space, and then the further residuals created by the quantized and unquantized latent space, described as a multi-stage vector quantizer in \cite{VQ_tech}. 
In practice, the \gls{rvq} is essential to deploy an efficient compressor at the edge; a single quantization layer would need a very large (indeed, exponentially large) codebook to represent the compressed data accurately. Conversely, \gls{rvq} offers an elegant solution. Indeed, it provides a progressively finer approximation to these high-dimensional vectors by employing a cascade of codebooks. The primary codebook offers a first-order quantization of the input vector. The residuals, or the differences between the data vectors and their quantized representations, are further quantized using a secondary codebook. Therefore, changing the number of codebooks affects the \gls{cr} and the data compression quality.

The codebook size and the codeword are programmable, as well as the number of quantizers. Our model is based on the open source \textit{vector-quantize-pytorch} library. \textit{EdgeCodec} is utilizing four separate quantizers, each equipped with 768 codewords, summing to about \qty{1.2}{\mega B}. The \gls{rvq} allows for variable bitrate by selecting a subset of quantizers. Consequently, we have four separate supported bitrates.

\subsubsection{Decoder}

\begin{figure*}[t]
    \centering
    \includegraphics[width=1\linewidth]{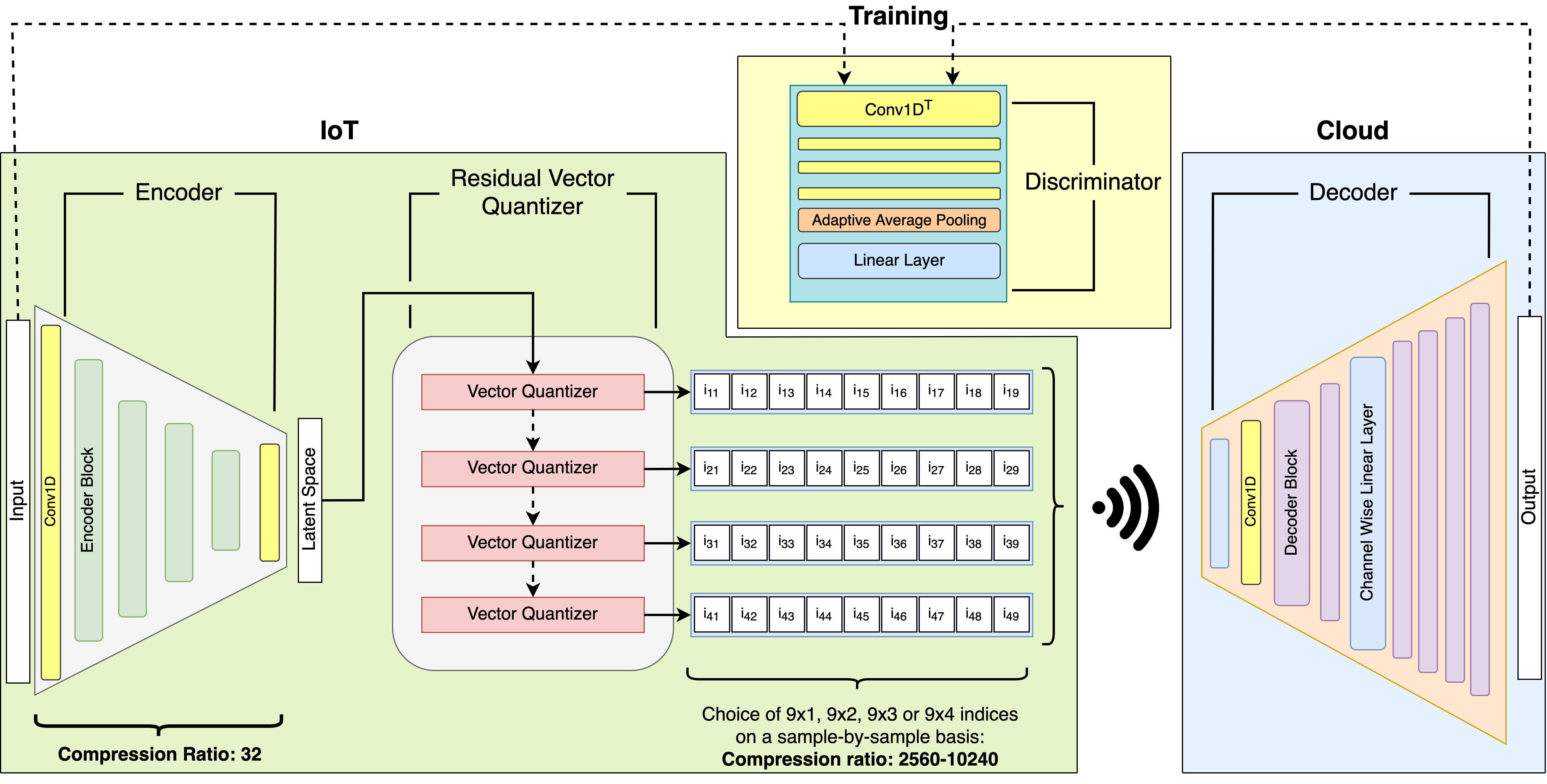}
    \caption{Detailed setup of \textit{EdgeCodec}. The model itself is subdivided into three parts in this figure. The one highlighted with a green background is the \gls{iot}-side where encoder and \gls{rvq} are deployed. The decoder, highlighted by the blue square, is on the cloud side. Moreover, there is a discriminator which is enabled only during training. The encoder has a native \gls{cr} of 32, which is further enhanced by the \gls{rvq} to be between 2560-10240 depending on number of quantizers used.}
    \label{fig:Detailed_model}
\end{figure*}

The decoder seen in \Cref{fig:Detailed_model} consists of a channel-wise linear layer, followed by a 1D-convolutional layer, afterwards two decoder blocks. Another channel-wise linear layer connects the second cascade of four decoder blocks, rounded off with a 1D-convolutional layer. Every decoder block consists of one transposed 1D-convolutional layer and three residual units. In this case, we use ELU as activation function. The decoder is designed to work on the cloud; therefore, \textit{EdgeCodecs} architecture is heavily unbalanced to shift the workload toward the server side while removing computational load from the edge. With the current setup, the decoder has a parameter count of 3’133’184, or \qty{12.5}{\mega B}. As the encoder so is the decoder designed in a compartmentalized fashion, making it easy to add or reduce complexity. For example, if local decompression is needed, one can make the decoder lighter.
\vspace{-5pt}
\subsection{Training and Evaluation}

The training was done end-to-end, with all three main blocks trained simultaneously. Moreover, a random number of quantizers is used, forcing the decoder to correctly perform with a variable \gls{cr}. Similarly to \cite{soundstream}, a composite loss function was used consisting of four separate losses $\mathcal{L}(\mathbf{x},\mathbf{\hat{x}}, \mathbf{z}, \mathbf{\tilde z}, \mathbf{y}, \mathbf{g}) = \alpha \cdot \mathcal{L}_{MSE}(\mathbf{x}, \mathbf{\hat{x}}) + (1{-}\alpha)\cdot \mathcal{L}_{l1\;smooth}(\mathbf{x}, \mathbf{\hat{x}})\quad+\eta \cdot \mathcal{L}_{rvq}(\mathbf{z}, \mathbf{\tilde z})+\gamma \cdot \mathcal{L}_{adv}(\mathbf{y}, \mathbf{g})$.
$\mathcal{L}_{MSE}(\mathbf{x}, \mathbf{\hat{x}})$ is a common \gls{mse}, which compares the input $\mathbf{x}$ and output $\mathbf{\hat{x}}$ of the model. It contributes to the decrease of the average difference between $\mathbf{x}$ and $\mathbf{\hat{x}}$. $\mathcal{L}_{l1\;smooth}(\mathbf{x}, \mathbf{\hat{x}})$ is a huber loss, which judges the input $\mathbf{x}$ and output $\mathbf{\hat{x}}$ of the model; it helps in uniforming the reconstructed signal in the presence of a high derivative gradient. $\mathcal{L}_{rvq}(\mathbf{z}, \mathbf{\tilde z})$ is also a \gls{mse}, but it judges the output of the encoder Enc($\mathbf{x}$) = $\mathbf{z}$ and the quantized latent space $\mathbf{\tilde z}$. Lastly, $\mathcal{L}_{adv}(\mathbf{y}, \mathbf{g})$ is the adversarial loss introduced by the discriminator. It is a binary cross-entropy-with-logits loss, where $\mathbf{y}$ denotes a logits vector full of zeros if we are judging the reconstructed output $\mathbf{\hat{x}}$ and full of ones if we are judging the original input $\mathbf{x}$. $\mathbf{g}$ is the respective output of the discriminator D, so $\mathbf{g} = D(\mathbf{p})$. The \textit{Adam} optimizer was used in training.

The evaluation was done by computing the average error per channel on the inmost 512 datapoints, due to the processing of the data by \textit{AeroSense} are the edges only padding. The evaluation itself is a percentage error of reconstructed data $\mathbf{\hat{x}}$ to original data $\mathbf{x}$ and then averaging it per channel.

\subsection{\gls{rvq} at the Edge}

For the scope of this work, we develop a \gls{rvq} and a parallelized \gls{vq} library, named p-\gls{vq}, in C code, which is compatible with most of the common RISC-V and ARM Cortex-M \gls{mcu}. In detail, \gls{rvq} works by cascading $N$ quantizers $Q_i$. At first, the latent space gets quantized in the first quantizer by calculating the Euclidean distance between the input $\mathbf{x}$ and the $k$ codewords $\mathbf{y_i}$, choosing the codeword with the lowest Euclidean distance to represent that input vector $\mathbf{x}$. Then a residual $\mathbf{\hat{x}_1}$ is calculated, which is $\mathbf{\hat{x}_1} = \mathbf{x}-\mathbf{y_{1i}}$, where $\mathbf{y_{1i}}$ is the i-th codeword of the first quantizer, and this residual is quantized by the second quantizer. In essence, the n-th residual produced by the n-the quantizer is defined as $\mathbf{\hat{x}_n} = \mathbf{x} - \sum_{i=0}^{n-1}\mathbf{\hat{x}_i}$. In \Cref{alg:gen_rvq}, we can see the pseudo-code for \gls{rvq}. We trained the \gls{rvq} to have bitstream generation flexibility and minimize the memory footprint. Moreover, a maximum reconstruction error of 3\%, motivated in \Cref{sec:dataset} is considered.

\begin{algorithm} 
    \small
    \LinesNotNumbered
    \KwIn{Input $x$ (Output of the encoder, vector quantizers $Q_i$, for $i = 1..N$}
    \KwOut{quantized $\hat{x}$}
    $\hat{x} = 0.0$\\
    $residual = x$\\
    \For{q = 1 to N}
    {
        $\hat{x} += Q_q(residual)$\\
        $residual -= Q_q(residual)$\\
    }
    \textbf{return} $\hat{x}$\\
\caption{\gls{rvq} pseudo-code}
\label{alg:gen_rvq}
\end{algorithm}
The thought process behind quantizing residuals is that we can extract more intricate details from our data. However, we are not restricted to always using all quantizers and their codewords. This algorithm can be stopped at any point, creating $N$ different bitrate modes. This enables a sample-by-sample chosen bitrate. The high \gls{cr} arises because only one index is transmitted per quantizer, making it possible to represent a whole channel with a single number. 
Assuming we compress numbers with precision $A$, and having $B$ of these numbers, the codeword size of $k$ is defined for each codebook. Therefore, the minimum number of bits representing the indices is calculated in \Cref{eq:general_bits}, from which the \gls{cr} is derived.
\begin{equation}
    \#bits = \lceil log_2(k)\rceil~,~CR = \frac{A \cdot B}{\#bits \cdot N}~.
    \label{eq:general_bits}
\end{equation}

%
\section{Results and Discussion}
\label{sec:Results}
\textit{EdgeCodec} is experimentally validated and deployed at the edge in the use case described in \Cref{sec:casestudy} using 32-bit wide samples generated by an array of 40 barometers sampled at \qty{100}{\hertz}. We employ a \gls{soa} \gls{mcu}, namely GAP9 from Greenwaves. GAP9 is a multi-core ultra-low power \gls{mcu} designed to run \gls{ML} at the edge. It features \qty{1.5}{MB} of onboard RAM (which entirely fits our \textit{EdgeCodec}) and an external non-volatile memory of \qty{64}{MB} to store the RVQ codebooks. The \textit{EdgeCodec} encoder block is optimized to run on GAP9, and the full pre-deployment validation is conducted with the NNTool from Greenwaves. A \textit{float16} representation is used for the implementation at the edge to achieve compression accuracy and optimize memory usage. At \qty{370}{\mega\hertz}, the measured GAP9 average power consumption is \qty{153}{\milli\watt}. 
\Cref{tab:mod_CR} shows the models \gls{cr}, bitrate, and average error over the whole validation set, segmented by how many quantizers were used for reconstruction. While achieving compression ratios consistently above 2000, the generated bitrate is in the range of a few bits per second and can be changed in real-time using a variable number of quantizers. 
\begin{table}[t]
    \centering
        \caption{Variable compression ratio and bitrate. Average error calculated over the whole validation set.}
    \begin{tabular}{l|r|r|r|r}
        \hline\hline
         \#Quantizers & 1 & 2 & 3 & 4\\
         CR $\uparrow$ & 10240 & 5120 & 3413 & 2560\\
     
        Bitrate (bps) & 11.25& 22.5& 33.75& 45\\
      
        Average Error $\downarrow$   & 2.93\%& 2.62\%& 2.56\%& 2.54\%\\
        \hline\hline
    \end{tabular}
    \label{tab:mod_CR}
\end{table}
We see that \textit{EdgeCodec} only incurs an extra error of 0.4\% while increasing its \gls{cr} by a factor of 4, from 2560 to 10240, respectively. Hitting compression ratios of that magnitude enables the usage of very low power transmission protocols, or very large coverage, due to the resulting bitrate being always below \qty{50}{bps}. Notably, the reconstruction error between 3 to 4 quantizers is not relevant, making the variable bitrate attractive since the error penalty is minor. 
\Cref{tab:quant_eval_mat} summarizes the results of two non-neural compressors, \gls{sz} \cite{SZ} and \gls{zfp} \cite{lindstrom2024zfp}, used as a reference comparison for the scope of this paper. They are evaluated on the same validation set used for our \textit{EdgeCodec}. Notably, \gls{sz} features an impressive low error of 0.103\%, but it is capped at a \gls{cr} of 512. Therefore, \textit{EdgeCodec} exhibits an error $25\times$ higher but a compression factor between 5 and 20 greater, and 128 to 512 increase compared to \gls{zfp}.

\begin{table}[t]
\centering
    \caption{\gls{cr} and average error over the whole validation set of different \gls{soa} compression techniques from the literature. SZ was evaluated in three different modes, ABS (\_A), REL (\_R), and REL/ABS (\_RA), which use different error bounds \cite{SZ}.}
    \begin{tabular}{lrrrr}
        \hline\hline
         & \textbf{ZFP} & \textbf{SZ3\_A} & \textbf{SZ3\_R} & \textbf{SZ3\_RA}\\
         \cmidrule(l){2-5}
        Avg. Error $\downarrow$& 2.85\%  & \textcolor{teal}{0.103\%}  & \textcolor{teal}{0.103\%}  & 9.03\%  \\
        \gls{cr} $\uparrow$& 15-19  & 512  & 512   & 512  \\
        \hline\hline
    \end{tabular}
\label{tab:quant_eval_mat}
\vspace{-5pt}
\end{table}

The choice of the four separate quantizers, each having $768$ codewords, came from the hyperparameter tuning during training and the memory consideration of the GAP9, our chosen \gls{mcu}. We explored several codewords and quantizers of 512-1024 and 2-6, respectively. As a rule of thumb, the optimal codebook usage lies around $\sim$85\%, so to support the reconstruction quality without wasting memory space, with unnecessarily large codebooks only partially used in the latent space. Codebooks with near 100\% utilization lead to increasingly degenerate results, making them unusable. In our specific use case, the validation loss between 4 and 6 quantizers is similar. In comparison, with 2 quantizers, we note a validation loss increase of 30\% and an average error increase of 0.06\% (see \Cref{tab:mod_CR}). Therefore, we selected 4 quantizers with $768$ codewords, each with a vector length of 100. Hence, the required size for storage is \qty{1.2}{MB}.

The implementation of the \gls{rvq} algorithm on the GAP9 can be roughly broken down into three most important phases \begin{enumerate*}[label=(\roman*), font=\itshape] \item memory allocation, \item extraction and evaluation, \item and finally saving. \end{enumerate*} To avoid on-the-go memory allocation, we start with an allocation phase in which all needed memory is reserved beforehand, and then only in-place changes are made. This is important due to the delay introduced by accessing and freeing memory. 
In essence, the algorithm walks over the number of quantizers used, the number of channels the encoder produces, and lastly the number of codewords per codebook. The second phase extracts the needed codeword from the memory where the current codebook is loaded. Afterward, the Euclidean distance is calculated between the channel and the codeword. In the last phase, we save the best index for that specific channel for every quantizer, leading us to an output matrix of shape $\mathbb{R}^{\#quantizers \; \times \; \#channels}$, so with four quantizers $9 \times 4$ in our case. Due to using 768 codewords, instead of full precision for the indices, we can use $\#bits = \lceil log_2(768)\rceil = 10$. \\
In \Cref{alg:pvq}, we can see the parallelized \gls{vq} implementation, also provided by this paper. It parallelizes the stage of calculating the Euclidean distance between input and the codewords, going from \qty{51}{\milli\second}, regular \gls{vq}, to \qty{7.34}{\milli\second}. This yields a speedup of $6.85\times$.

\begin{algorithm}[t]
\small
\LinesNotNumbered
    \small
    \KwIn{Input $x$, quantizer $Q$}
    \KwOut{Codebook Indices $C$}
    \For{i = 0 to N\_CHANNELS}
    {
        $fork\_res = vec  \in \mathbb{R}^{N\_CHANNELS}$ \\
        $current\_distance = 10e6$ \\
        \underline{parallel} \For{k = START to END}
        {
        $cdist\_curr = \textbf{cdist(x, Q)}$\\
        $if\;cdist\_curr < current\_distance$\\
        \{\\
            $current\_distance = cdist\_curr$\\
            $k\_current = k$\\
        \}\\
        }
        $best\_result = \textbf{find\_best(fork\_res)}$\\
        $C[i] = best\_result$\\
    }
    \textbf{return} $C$
\caption{Pseudo-code for parallel \gls{vq}}
\label{alg:pvq}
\end{algorithm}

The execution time profiling of the cloud-side, including the \gls{rvq} and the decoder, is performed on an Intel(R) Core(TM) i7-10700K CPU running at \qty{3.8}{\giga\hertz} and an NVIDIA GeForce RTX 4070 Ti GPU. The \gls{mcu} profiling was done through GVSOC, a fully resource restricted simulator provided by \textit{GreenWaves Technologies} to simulate real GAP9 deployment. We can see the inference times summarized in \Cref{tab:MC_speed}. The Encoder block is executed in \qty{45.3}{\milli\second} using \textit{float16} precision, while the single core \gls{rvq} requires \qty{250}{\milli\second}, a time heavily dominated by memory transfer operations. Using the multi-core capabilities of GAP9, the execution time of the \gls{vq} is reduced to \qty{7.34}{\milli\second}, referred as p-\gls{vq} in \Cref{tab:MC_speed}. In \Cref{fig:Detailed_model}, this part is highlighted in the \gls{iot} block. Notably, the execution time is not always deterministic, as it changes with the \# of quantizers.

\begin{table}[t]
    \centering
    \caption{\textit{EdgeCodec} execution time and memory footprint in millisecond and MB, respectively. GAP9 is used as reference \gls{mcu} for the Encoder, p -VQ (parallelized - \gls{vq}), and \gls{rvq}, running on the sensor node. Input rebuild is the indexing of the codebooks and rebuilding the quantized latent space, which is deployed together with the Decode on the cloud side.}
    \begin{tabular}{lccccc}
        \hline\hline
         & \textbf{Encoder} & \textbf{p -VQ} & \textbf{RVQ} &\textbf{Input rebuild} & \textbf{Decoder} \\
        \hline
         Execution & 45.3 & 7.34 & 250 & 0.331 & 5.175 \\
         Memory  & 0.232 & 1.2 & 1.2 &  & 12.5 \\
        \hline\hline
    \end{tabular}
\label{tab:cloud_speed}
    \label{tab:MC_speed}
\end{table}

The AeroSense node gathers data with a frequency of \qty{100}{\hertz}; therefore, the compression pipeline needs to be faster than \qty{8}{\second}, given the \textit{EdgeCodec} input windows of 800 samples. In \Cref{tab:MC_speed}, we can see that the \gls{mcu} execution time is largely below half a second, if utilizing the \gls{rvq}, and well below a tenth of a second if using p-\gls{vq}. In \Cref{tab:cloud_speed}, we can see that the reconstruction and decompression takes roughly \qty{5.5}{\milli\second}, making the whole pipeline run in \qty{300}{\milli\second} for a single measurement, which makes the whole pipeline real-time capable using only 4\% of the time budget.

In \Cref{tab:MC_speed}, we can see an overview of the memory footprint. The most important parts are the encoder and \gls{rvq} since they have to fit into the limited memory of the GAP9. With the encoder being \qty{232}{\kilo B} and all four codebooks coming to \qty{1.2}{\mega B}, both fit into the onboard memory block of the chosen \gls{mcu}.

The 40 barometers used as a case study for this paper generate a bitrate of \qty{128}{kbps}, as calculated in \Cref{eq:AeroSense_bitrate}. 
\begin{equation}
    bitrate=\frac{40\cdot800\cdot32bit}{8s}=128kbps
\label{eq:AeroSense_bitrate}
\end{equation}
Over \gls{BLE}, the data transfer needs an energy of \qty{23}{\milli\joule}. Minimizing the energy consumption is crucial for real-world deployment since battery-supplied \gls{iot} sensor nodes have limited energy availability. Deploying \textit{EdgeCodec} would reduce the bitrate by a factor of $2560-10240\times$. As shown in \Cref{tab:cloud_speed}, the execution time at the edge settles down to $Time_{Edge}=Encoder+pVQ =$\qty{52.6}{\milli\second}, with an equivalent energy of \qty{8}{\milli\joule}. Considering the best case with the highest compression factor, the transmission power is reduced to \qty{9}{\micro\joule}. Therefore, \textit{EdgeCodec} can save up to $2.9\times$ of energy required for wireless data transmission.
\section{Conclusions}

This paper proposes \textit{EdgeCodec}, an end-to-end trained neural compressor, which utilizes the novel compression technique \gls{rvq} and an adversarial training environment. Working on real in-the-field collected data by \textit{AeroSense}, which are compressed with an average error of 2.54-2.93\%, this being below the preset threshold described in \ref{sec:dataset}, and a \gls{cr} of 2560-10240. The whole pipeline supports a sample-by-sample variable bitrate and runs entirely in real time. \textit{EdgeCodec} can address the problem of modern data collection and trafficking by \gls{iot}-devices. Moreover, it is a proof of concept and viability of adapting neural compressors to a domain where time-series data dominates the traffic. 



\bibliographystyle{IEEEtran}
\bibliography{bibliography}

\end{document}